\batchmode
\makeatletter
\def\input@path{{/Users/neil/Documents/neil/Research/ThingTank/pubs/2014.Turing-Test_Internet-of-Things//}}
\makeatother
\documentclass[english]{IEEEtran}
\usepackage[T1]{fontenc}
\usepackage[latin9]{inputenc}
\usepackage{refstyle}
\usepackage{graphicx}
\PassOptionsToPackage{normalem}{ulem}
\usepackage{ulem}

\makeatletter

\AtBeginDocument{\providecommand\figref[1]{\ref{fig:#1}}}
\AtBeginDocument{\providecommand\secref[1]{\ref{sec:#1}}}
\RS@ifundefined{subref}
  {\def\RSsubtxt{section~}\newref{sub}{name = \RSsubtxt}}
  {}
\RS@ifundefined{thmref}
  {\def\RSthmtxt{theorem~}\newref{thm}{name = \RSthmtxt}}
  {}
\RS@ifundefined{lemref}
  {\def\RSlemtxt{lemma~}\newref{lem}{name = \RSlemtxt}}
  {}

\newcommand{\lyxaddress}[1]{
\par {\raggedright #1
\vspace{1.4em}
\noindent\par}
}

\makeatother

\usepackage{babel}
\usepackage{listings}

\begin{document}

\title{Turing Test for the Internet of Things}

\author{Neil Rubens $^{1,2,3}$ }

\maketitle

\lyxaddress{$^{1}$ Active Intelligence Lab\\
$^{2}$ Skolkovo Institute of Science and Technology\\
$^{3}$ University of Electro-Communications}
\begin{abstract}
Draft: Work in Progress:

How smart is your kettle? How smart are things in your kitchen, your
house, your neighborhood, on the internet? With the advent of Internet
of Things, and the move of making devices `smart' by utilizing AI,
a natural question arrises, how can we evaluate the progress. The
standard way of evaluating AI is through the Turing Test. While Turing
Test was designed for AI; the device that it was tailored to was a
computer. Applying the test to variety of devices that constitute
Internet of Things poses a number of challenges which could be addressed
through a number of adaptations.
\end{abstract}

\section{Introduction}

With the advent of the Internet of Things (IOT), and the move of making
devices ``smart'' by utilizing AI \cite{Kortuem2010}, a natural
question arrises: ``How can we evaluate the progress of making devices
smarter?''. A common approach of evaluating AI is through the Turing
Test (TT) \cite{Turing1950a} in which interrogator through a text
conversation with two participants tries to determine which is a computer
and which is a human (\figref{Turing-Test-Cartoon}). However, in
the context of the Internet of Things, Turing test (in its traditional
form) may not be directly applicable in part due to the following:
\begin{enumerate}
\item Turing Test assumes that AI is embodied in a \textbf{single computer}:

\begin{quote}
``The present interest in `thinking machines' has been aroused by
a particular kind of machine, usually called an `electronic computer'
or `digital computer'. Following this suggestion \uline{we only
permit digital computers to take part in our game}.'' \cite{Turing1950a}\end{quote}
\begin{itemize}
\item Internet of Things

\begin{itemize}
\item multiple (connected) devices
\item many of the devices contain ``computers'', but are not computers
per-say (e.g. kettle or fridge).
\end{itemize}
\end{itemize}
\item Turing Test focuses on measuring \textbf{conversational intelligence}:

\begin{quote}
``The ideal arrangement is to have a \uline{teleprinter communicating
between the two rooms}.'' \cite{Turing1950a}\end{quote}
\begin{itemize}
\item Internet of Things

\begin{itemize}
\item some of the devices have no means of conducting a conversation (e.g.
a toaster).
\end{itemize}
\end{itemize}
\end{enumerate}
Evaluating AI in the context of IOT seems to require some adaptations,
descriptions of which we provide in this paper. Similarly to the original
Turning test paper \cite{Turing1950a}; we avoid defining ``intelligence''
and instead concentrate on practical/pragmatic aspects.
\begin{quote}
\begin{figure}
\includegraphics[width=1\columnwidth]{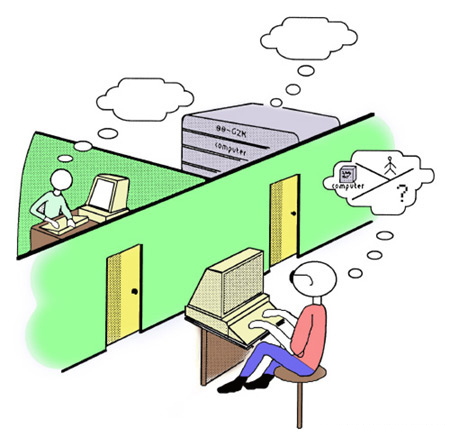}

\protect\caption{Turing Test: interrogator through a text conversation with two participants
tries to determine which is a computer and which is a human \cite{Copeland:1993:AIP:530166}.\label{fig:Turing-Test-Cartoon}}

\end{figure}

\end{quote}

\section{Simplified Turing Test for the Internet of Things}

\begin{figure}
\begin{centering}
\includegraphics[width=0.35\columnwidth]{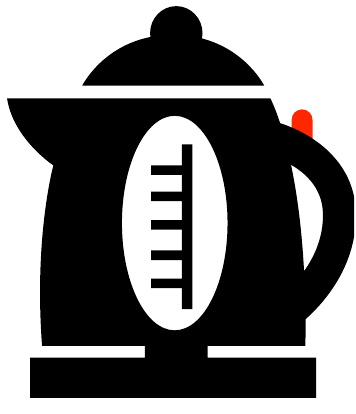}
\par\end{centering}

\protect\caption{Electric Kettle.\label{fig:kettle}}
\end{figure}

Let us propose a straightforward adaption of the TT to IOT; followed
by describing the challenges that the test would face (in \secref{proposal}
we propose extensions to TT that allow to alleviate at least some
of the challenges). In this version we strive to make as few modifications
to the original turing test as possible. Hence, we only modify the
medium through which the ``conversation'' happens by replacing computer
terminal with a ``thing'' (e.g. an electric kettle (\figref{kettle})).
To add the i/o capabilities, we add a remote input/control to the
kettle (that controls the on/off switch) (similar to the keyboard
in the original test). Interrogator is also able to interact with
``kettle'' through ``local'' on/off switch.

the communication channel (sending on/off status along with timestamps).
One of the kettles would be controlled by a person and the other one
would be controlled by a computer; interrogator's task would be to
decide which is which.

\section{Challenges for Turning Test in the Context of the Internet of Things\label{sec:challenges}}

Turing test has been a truly visionary work that became a standard
for evaluating performance of AI systems; preceding by 6 years the
official establishment of the field of Artificial Intelligence \cite{BJH:813}.
However, in 1950 it was difficult to foresee all of upcoming developments
which would affect the development of AI. In this section, we describe
various aspects of the Internet of Things that make it difficult to
apply TT in a straightforward manner.

\subsection{Limited Interface\label{sec:Limited-Interface}}

In Turning test interaction is assumed to happen in the text conversation
format. Many of the things constituting IOT have limited interfaces;
e.g. a kettle has only on/off switch.

\subsection{Limited Computational Capabilities\label{sec:Limited-Computation}}

Many of the IOT devices have limited computing capabilities which
are often geared to very specific purposes (e.g. auto-shutoff controller
of the kettle).

\subsection{Intelligent Behavior that is not Human Behavior\label{sec:Intelligent-Human-Behavior-Challenge}}

\begin{figure}
\includegraphics[width=1\columnwidth]{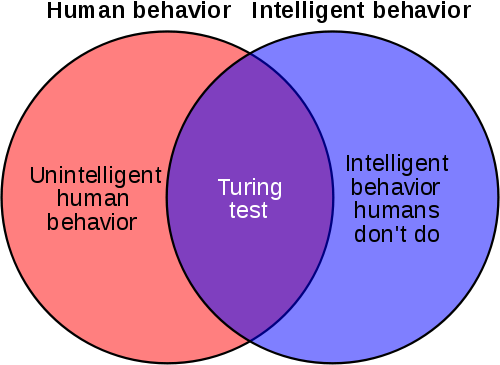}

\protect\caption{Turing Test: Intelligent Behavior vs Human Behavior \cite{wiki:turing:test}.}
\label{fig:intelligent-behavior}
\end{figure}

The Turing test is a test of a machine's ability to exhibit intelligent
behavior indistinguishable from that of a human. However, one of the
major drawbacks of Turing Test is that not all of the intelligent
behavior is human-like (\figref{intelligent-behavior}). This is exacerbated
if we consider that the test medium is not a textual conversation
(which many things are not capable of functionally \secref{Limited-Interface}).
Bellow we outline some of the challenges that are exharbated in the
context of IOT.

\subsubsection{Analysis of Large Quantities of Numerical Data\label{sec:Analysis-Challenge}}

Most people may have difficulty with analyzing large quantities of
numerical data (especially without analytical tools); more so if it
comes from multiple sources.

\subsubsection{Response Time\label{sec:response-time-challenge}}

People are not able to respond as quickly as machines.

\subsubsection{Time: Availability / Duration}

Turning test assumes a relatively short pre-scheduled session during
which examiner performs a test: ``mating the right identification
after five minutes of questioning'' \cite{Turing1950a}. The natural
test duration for testing of things could be much longer; e.g. examining
if your coffee making machine is intelligent could take weeks or months;
and not necessarily at pre-scheduled times (e.g. if you happen to
wake earlier than usual and need a sip of coffee).

\subsubsection{Memory / History}

In the original description it wasn't stated explicitly if there is
a conversational memory (script).

and we also add the conversation ``memory'' function that shows
history of on/off activity (details ...); similar to the text conversation
history displayed on the screen.

for this settings peoples memory is more limited than that of a computer

\subsection{Evaluation Criteria\label{sec:Evaluation-Criteria-Challenge}}

Turing test evaluates conversational intelligence. However it is not
clear wether this criteria makes sense in different context; e.g.
is it appropriate to measure how ``intelligent'' a kettle is by
having a conversation with it; what if it talks well, but doesn't
perform well as a kettle (starts pondering the meaning of life, while
deciding wether to boil the water)?

\subsection{Multiple Participants}

Turning test assumes interaction with a single participant at a time
(be it a human or a computer). Internet of Things as the name implies,
assumes that there are multiple entities involved. Therefore the challenge
is adopting Turing test to multiple participants. 

\begin{figure*}
\begin{centering}
\begin{lstlisting}
I: day1: 8:00am : turn on kettle
T: day1: 8:02am : turn off kettle
I: day1: 8:10am : turn on kettle
T: day1: 8:12am : turn off kettle

T: day2: 7:58am : turn on kettle
T: day2: 8:00am : turn off kettle
I: day2: 8:10am : turn on kettle
T: day2: 8:12am : turn off kettle

T: day3: 8:09am : turn on kettle
T: day3: 8:12am : turn off kettle
\end{lstlisting}

\par\end{centering}

\protect\caption{Interaction Intelligence Criteria. I - interrogator, T - participating
thing (controlled by computer or a human) see \secref{Interaction-Intelligence}.\label{fig:dialogue-script}}
\end{figure*}

\section{Proposed Approaches\label{sec:proposal}}

In this section we describe approaches to addressing challenges of
adopting the Turing Test to the context of the Internet of Things
(\ref{sec:challenges}).

\subsection{Limited Interface}

As described in (\secref{Limited-Interface}) many of the things constituting
IOT have limited interfaces; e.g. a kettle has only on/off switch.
One might suggest that the original interface (e.g. on/off switch)
could be used to do a binary encoding of a text conversation. Another
option is to add an interface for a text conversation. However, even
once interface allows for the textual conversation; other related
challenges still remain unanswered: would the device computationally
capable of having an intelligent conversation (\secref{Limited-Computation}),
and is conversation the proper way to evaluate the intelligence of
a device (\secref{Evaluation-Criteria-Challenge}). Answer to these
questions might dictate the form of the interface. Our suggestion
is to leave the interface as is.

\subsection{Limited Computational Capabilities\label{sec:Limited-Computation-Proposal}}

As described in (\secref{Limited-Computation}) many of the devices
have limited computational power and/or target for specific purposes
(e.g. micro-controller in kettle for the auto-shut-off function).
It is possible to modify/add hardware and software as to allow for
generating and processing of textual conversations (either locally
or remotely). However in this case, we might be appraising the intelligence
of the add-ons rather than that of the device (which is our original
goal). Hence our suggestion is to leave the hardware and software
as is.

\subsection{Intelligent Behavior that is not Human Behavior}

While previous challenges were favorable to humans; the challenges
in this section favors computers; since we focus on the intelligent
behavior that humans don't do (\secref{Intelligent-Human-Behavior-Challenge}).
We do want to make this a fair test hence we try to address these
challenges. In the original paper \cite{Turing1950a} this challenge
was pointed out:
\begin{quote}
``It is claimed that the interrogator could distinguish the machine
from the man simply by setting them a number of problems in arithmetic.
The machine would be unmasked because of its deadly accuracy.'' 
\end{quote}
In the same paper \cite{Turing1950a} a potential solution was suggested:
\begin{quote}
``The machine (programmed for playing the game) would not attempt
to give the right answers to the arithmetic problems. It would deliberately
introduce mistakes in a manner calculated to confuse the interrogator.''
\end{quote}
Making AI appear dumber and slower than it really is, has been used
in programs aimed at passing the turing test, as well as in industry
e.g. gaming \cite{liden2003artificial}. 

``Dumbing down'' approach could certainly be introduced in the context
of IOT in a fairly straightforward manner. However, we strongly feel
that the aim should be to achieve a high intelligence, rather than
to simply mimic human behavior (\secref{Evaluation-Criteria-Approach}).
Hence we take a different approach of attempting to make human participants
to appear more ``computationally intelligent''. For the original
test this could be achieved be equipping participants with calculators.
In the context of IOT, additional modifications may be needed. To
narrow the gap between human and computers (for the tasks in which
computers perform better); we consider two non-exclusive approaches:
human-centered and computer assisted.

\subsubsection{Human-centered}

A typical approach of improving processing speed of computers is through: 
\begin{itemize}
\item faster processing
\item parallelization
\end{itemize}
People's response time could be improved in a similar manner:
\begin{itemize}
\item faster processing

\begin{itemize}
\item using more skilled human operators
\item training human operators
\end{itemize}
\item parallelization

\begin{itemize}
\item using multiple participants
\end{itemize}
\end{itemize}
This approach does narrow the performance gap but not sufficient,
and it is ``expensive''. If this approach is used, it should also
be complemented with the computer-assisted approach.

\subsubsection{Computer-assisted\label{sec:Computer-assisted}}

Receiving assistance from computers should allow to narrow (if not
exceed) some of the performance gaps. In particular computers could
be used to assist with: (1) analysis, (2) execution).

\paragraph{Computer-assisted Analysis}

Using analytic software would allow to address the challenge of analyzing
large quantities of numerical data (\secref{Analysis-Challenge}).

\paragraph{Computer-assisted Execution}

The rest of the challenges (\secref{Intelligent-Human-Behavior-Challenge})
could be addressed by programming computers with the desired responses
or analysis algorithms. In order to keep the distinction between pure
AI and computer-assisted approach people should have ability to override
computer's decisions.

\subsection{Evaluation Criteria\label{sec:Evaluation-Criteria-Approach}}

Turing test evaluates conversational intelligence. However it is not
clear wether this criteria makes sense in different context; e.g.
is it appropriate to measure how ``intelligent'' a kettle is by
having a conversation with it; what if it talks well, but doesn't
perform well as a kettle (starts pondering the meaning of life for
an hour, while deciding wether to boil the water)? On the other hand,
focusing on evaluation operational performance does not necessarily
reflects how ``intelligent'' the thing is. Bellow we propose some
of the approaches that could be used (in combination or independently).

\subsubsection{Interaction Intelligence \label{sec:Interaction-Intelligence}}

In order to overcome some of the operational speed capabilities of
humans and computers (\secref{Intelligent-Human-Behavior-Challenge});
we can evaluate intelligence through a form of dialogue. However,
the dialogue is not be textual but is operational (\figref{dialogue-script}).
Both the interrogator and participants can control the time-stamps
as to simulate the flow of time). Note that interrogator can prepare
multiple dialogue scripts in order to examine the responses of the
participants. In the end; based on the responses interrogator would
have to decide wether the participant (interrogated thing) is controlled
by a human or a computer.

\subsubsection{Operational Intelligence\label{sec:Operational-Intelligence}}

Devices could be evaluated based on how ``intelligently'' they behave.
Unlike pure operational evaluation (e.g. how quickly the water is
boiled); this criteria would also consider intelligence aspects, e.g.
is a kettle able to distinguish between different users (members of
the family), different usages (boiling water for tea, or for drip
coffee, or other purposes).

\subsubsection{Inner Intelligence}

While the operational intelligence reflects the intelligence of the
behavior of the objects (\secref{Operational-Intelligence}); inner
intelligence evaluation allows users to examine the internal rules/workings
of the things to determine if those are intelligent or not. However
this evaluation assumes that computer-assisted execution method was
used (\secref{Computer-assisted}), and that resulting rules could
be represented in a human readable form e.g. if then rules or decision
trees; however other methods are very hard to either create or examine
manually e.g. SVM, random forests, neural networks.

\subsection{Multiple Participants}

Turning test assumes interaction with a single participant at a time
(be it a human or a computer). Internet of Things, as the name implies,
assumes that there are multiple entities involved. Therefore the challenge
is adopting Turing test to multiple participants. 

Assuming multiple participants raises more questions about evaluating:
\begin{itemize}
\item collective/distributed intelligence 
\item mixed groups

\begin{itemize}
\item different devices
\item different types of entities (humans \& non-humans)
\end{itemize}
\item multiple evaluators
\item many-many or one-many (e.g. one-many: one person controlling multiple
things)
\item participating vs non-participating entities (in the evaluation)
\end{itemize}
Naive approach would be to simply apply the single-version of the
test to multiple objects and aggregate the results. More interesting
approaches could be obtained by being somewhat creative; e.g. mixing
human-controlled things with computer-controlled, etc.

\section{Conclusion}

It's been over 50 years since the Turing Test for evaluating AI has
been proposed. TT has stood up very well to the test of times; and
it is still by far the most prevalent way in which AI is evaluated.
However, new settings, in particular the Internet of Things, pose
challenges to the traditional definition of the Turing Test (\secref{challenges}).
We show that with some modifications (\secref{proposal}) the Turing
Test could be adapted to these novel settings.

\section*{Acknowledgement}

This research is performed in conjunction with the ThingTank project
\cite{TT:Anthropology:2014} a part of the Ideas Lab MIT-Skoltech
Initiative.

\bibliographystyle{abbrv}
\bibliography{3_Users_neil_Documents_neil_Research_bib_mendeley_library,4_Users_neil_Documents_neil_Research_bib_neil,5_Users_neil_Documents_neil_Research_bib_rubens}

\end{document}